\documentclass{article}

\usepackage{microtype}
\usepackage{graphicx}
\usepackage{subfigure}
\usepackage{booktabs} 
\usepackage{todonotes}
\usepackage{amsfonts}
\usepackage{amsmath}
\usepackage{hyperref}

\usepackage[accepted]{icml2021}

\icmltitlerunning{On the overlooked issue of defining explanation objectives for local-surrogate explainers}

\begin{document}

\twocolumn[
\icmltitle{On the overlooked issue of defining explanation objectives for local-surrogate explainers}




\icmlsetsymbol{equal}{*}

\begin{icmlauthorlist}
\icmlauthor{Rafael Poyiadzi}{equal,to}
\icmlauthor{Xavier Renard}{equal,axa}
\icmlauthor{Thibault Laugel}{axa}
\icmlauthor{Raul Santos-Rodriguez}{to}
\icmlauthor{Marcin Detyniecki}{axa,lip6,pol}
\end{icmlauthorlist}

\icmlaffiliation{to}{Department of Engineering Mathematics, University of Bristol, Bristol, United Kingdom}
\icmlaffiliation{axa}{AXA, Paris, France}
\icmlaffiliation{lip6}{Sorbonne Université, CNRS, LIP6, F-75005, Paris, France}
\icmlaffiliation{pol}{Polish Academy of
Science, IBS PAN, Warsaw, Poland}

\icmlcorrespondingauthor{Rafael Poyiadzi}{rp13102@bristol.ac.uk}
\icmlcorrespondingauthor{Xavier Renard}{xavier.renard@axa.com}
\icmlcorrespondingauthor{Thibault Laugel}{thibault.laugel@axa.com}

\icmlkeywords{Machine Learning, ICML}

\vskip 0.3in
]



\printAffiliationsAndNotice{\icmlEqualContribution} 

\begin{abstract}
Local surrogate approaches for explaining machine learning model predictions have appealing properties, such as being model-agnostic and flexible in their modelling. Several methods exist that fit this description and share this goal. However, despite their shared overall procedure, they set out different objectives, extract different information from the black-box, and consequently produce diverse explanations, that are -in general- incomparable. In this work we review the similarities and differences amongst multiple methods, with a particular focus on what information they extract from the model, as this has large impact on the output: the explanation. We discuss the implications of the lack of agreement, and clarity, amongst the methods' objectives on the research and practice of explainability.
\end{abstract}

\section{Introduction}

The need for machine learning interpretability, for example to understand a specific prediction, may arise for multiple reasons such as regulation, ethics, business requirements or model conception and control. 
From this multiplicity comes a variety of interpretability specifications.
For instance, let us consider a situation where a customer is denied a credit by a model.
The customer may want explanations for the differences between their situation and the ones of similar customers with an accepted application.
On the other hand, for the same model, a regulator may be looking for explanations to identify potential discrimination threats.
Because these stakeholders have different needs and goals, they most likely require different information, which should then be formalized into different mathematical objectives. 
Therefore, a careful analysis of the use-case should be carried out to define the interpretability objective of each situation and choose the most appropriate interpretability method, as it is unlikely that a one-size-fits-all solution exists. 


Yet, we argue that the current literature on model surrogates to explain a prediction lacks the clarity needed for a practitioner to make an informed choice on which method to use, given explanation needs.
Existing approaches usually lack transparency with regards to the explanation needs they propose to solve, on their specifications and ultimately on their formal objectives.
This situation (1) fuels a disseminated research with propositions that are difficult to compare and (2) prevents a sound development of the explainability practice.

In this paper, we propose a study to highlight the diversity amongst the approaches categorized under the same vague objective of ``explaining a prediction with a model surrogate''.
This work is based on a theoretical analysis of proposed solutions and an experiment to illustrate the differences.
In view of its popularity, an emphasis is given to KernelSHAP \cite{Lundberg_Lee_2017}.
Finally, we propose a discussion on the ways forward.
We first remind the construction principles shared across model surrogates to explain a prediction.
\section{Explaining Predictions with Surrogates}
\label{sec:background}

A surrogate in our context is a simple \emph{interpretable} machine learning model (e.g. linear model, decision tree or rules) that aims to mimic the predictive behaviour of a black-box model that is to explain.
As the surrogate model is interpretable, it needs to be simpler than the black-box model.
To preserve the fidelity of the model surrogate to the black-box despite its simplicity, the information the surrogate has to model needs to be restricted.
In particular, to explain a prediction, a surrogate will mimic the black-box only in the \textit{neighbourhood} of the instance whose prediction is to explain.
This defines the relevant subspace for explaining the prediction.



The general procedure to explain a prediction, made by a black-box classifier $f:\mathcal{X}\rightarrow\mathcal{Y}$ trained on a dataset $(\boldsymbol{X}, \boldsymbol{y}) = \{(\boldsymbol{x}_i, y_i)\}_{i=1}^n \in (\mathbb{R}^d\times \{0, 1\})$, with an interpretable model surrogate $g$ follows two steps, common to all approaches:

\begin{enumerate}
    \item Extract the relevant classification behaviour for the prediction of $\boldsymbol{z}_e$, by constructing the neighbourhood dataset $\mathcal{D}_N$, which describes the relevant predictive behaviour of the black-box $f$ for the prediction to explain. Optionally assign weights~$\boldsymbol{w}_{\boldsymbol{z}_e}$ to the instances of the generated neighbourhood.
    \item Train an interpretable model, the surrogate $g$, from an \emph{interpretable model class}, using the neighbourhood, and accompanying weights.
\end{enumerate}

The training procedure of the surrogate $g$ on the neighbourhood dataset $\mathcal{D}_N$ follows a standard machine learning procedure, apart from the trade-off fidelity-interpretability for $g$ as the model surrogate allows the direct generation of explanations.
However, the generation of the neighbourhood needs very specific attention as it directly controls the subspace where the surrogate mimics the black-box, and hence the \textbf{meaning} of the surrogate's explanations.
This issue is critical and the main challenge for making interpretable model surrogates comply with explainability specifications and formal objectives.

In the following section we analyse how interpretable model surrogates generate their neighbourhoods to highlight the diversity of views and the need for more clarity.

\section{Where Interpretable Surrogates Look: a Comparison of Neighbourhood Strategies}
\label{sec:comparison}

\begin{figure}[!t]
    \includegraphics[width=\columnwidth]{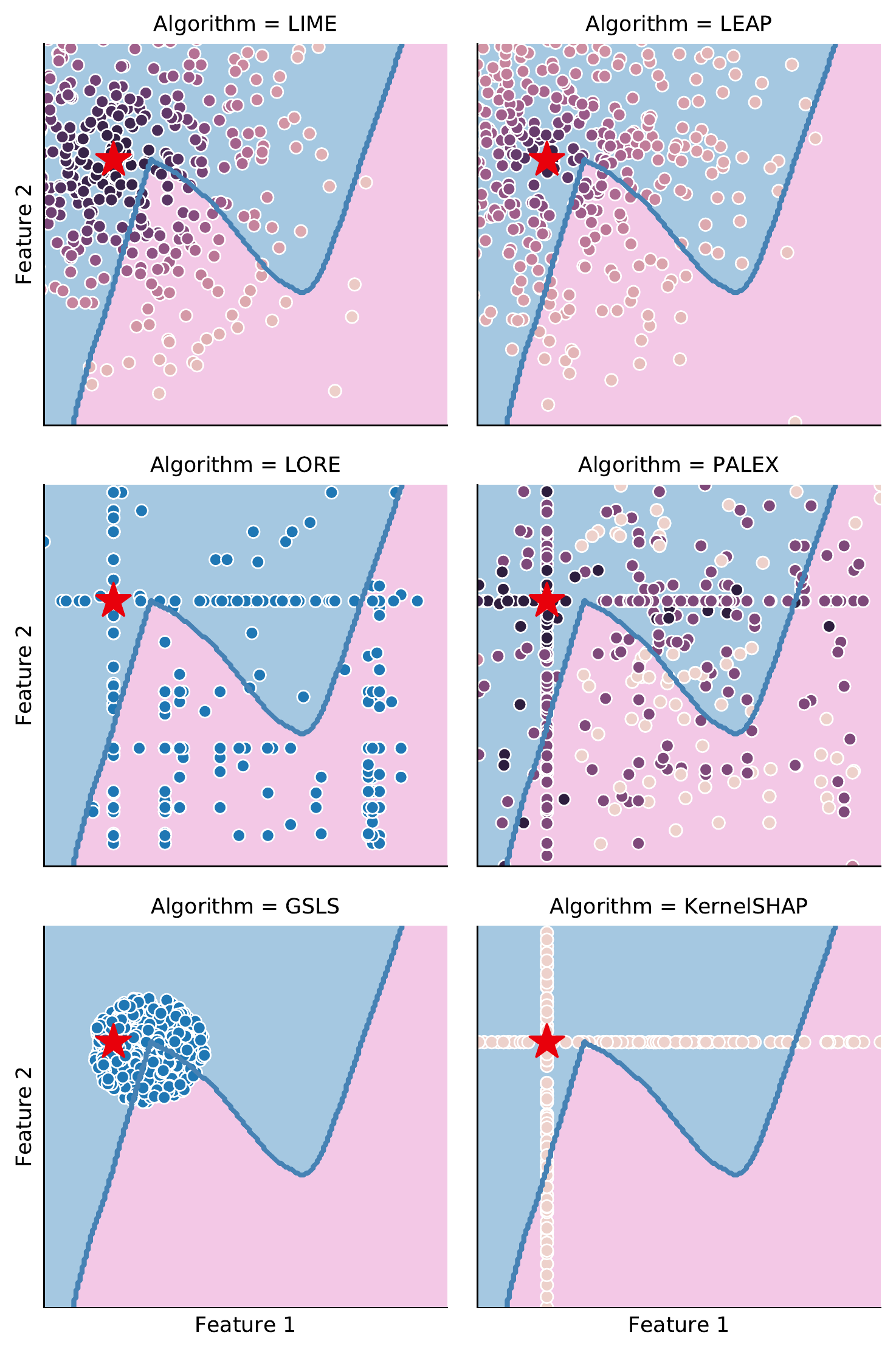}
    \caption{Neighbourhoods associated with the instances' weights (when applicable) generated by various local interpretable surrogate approaches. Setup: \textit{Half-Moons} dataset (red and blue points: 2 classes), neural network black-box represented by its decision boundary (red and blue areas). The instance of the prediction to explain (red star) is shared across the approaches. Points of the neighbourhoods are represented either with blue dots (no weights associated) or dots colored to represent the weight's value (darker equals more important weight). The neighborhoods generated, and thus the information captured to fit the surrogate, are very different while the setup and the prediction to explain are the same. }%
    \label{fig:sota_neighbourhoods}%
\end{figure}

\begin{table*}[t]
\centering
\caption{Comparison of the different neighbourhoods sampled by interpretable model surrogate approaches.}
\label{tab:comparison}
\resizebox{\textwidth}{!}{
\begin{tabular}{l||p{15cm}|p{7cm}|p{7cm}}
Algorithm & Method Description & Sampling & Weighting \\ \hline\hline
LIME
& Sampling from a Gaussian distribution centered on $z_e$ with points $x$ weighted for locality. 
& $x \sim \mathcal{N}(z_e, \sigma_{Lime})$
& $w_i = \exp\left(-\frac{1}{\gamma}\cdot \|x_i - z_e\|^2\right)$
\\\hline
GSLS
& Sampling from a Uniform distribution centered on $v^*$. $v^*$ is the nearest point from $z_e$ such that $f(v^*) \neq f(z_e)$.
& $x \sim \mathcal{U}(B_{v^*})$, $B_{v^*}=\{x \in \mathbb{R}^{d} :\|x - v^*\|_2 \leq r \}$
& - 
\\\hline
LEAP
& LIME procedure on a subspace $\mathcal{\hat{X}} \subseteq \mathcal{X}$, which is a product o local dimensionality reduction around $z_e$. $\hat{x}$ represents the projection of $x$ onto $\mathcal{\hat{X}}$.
& $x \sim \mathcal{N}(\hat{z}_e, \sigma_{Leap})$ with $\hat{z}_e \in  \mathcal{\hat{X}}$
& $w_i = \exp\left(-\frac{1}{\gamma}\cdot \|\hat{x}_i - \hat{z}_e\|^2\right)$
\\\hline
PALEX
& Repeatedly replaces values of a random subset of the features of $z_e$ with values from $\boldsymbol{X}$
& repeat:\newline
--~Sample random subset $Q$;\newline
--~$x_{\neg Q}$ drawn uniformly $\& z_{e, Q}$ copy from $z_e$;\newline
--~$x = [z_{e, Q}, x_{\neg Q}]$
& Co-appearance of $(z_e,x)$ in frequent patterns.
\\\hline
LORE
&  Tries to generate a neighbourhood,  where instances have feature
characteristics similar to the ones of $\boldsymbol{z}_e$, that is able to reproduce the local decision behavior of the black box.
&  Instances are generated with the use of a genetic algorithm.
& -
\\\hline
k-SHAP
& Formulates a weighted linear regression problem whose solution is the Shapley Values.
& \textit{for} (all subsets $Q \subset S$):\newline
--~$x_{Q} = 1~~\&~~x_{\neg Q} = 0$;\newline
--~$y_Q = \mathbb{E}[f(Z)~|~Z_i=z_{e,i},~\forall i \in Q]$
& Shapley kernel weights\newline
$k\left(x_{Q}\right)=\frac{d-1}{(d \text { choose } |s|) s(d-s)}$\newline
$s=|x_{Q}|$~the number of non-zero elements in $x_{Q}$.
\\
\end{tabular}
}
\end{table*}

In the previous section we discussed how the approaches have common construction process and made apparent the similarities between them.
In this section we analyse interpretable surrogate approaches and focus in particular on what information each approach extracts from the black-box: their neighbourhoods.
This analysis shows that beyond the common construction process, approaches differ in what information they extract from the black-box, and therefore produce different explanations.



Arguably one of the most well-known local-surrogate approaches is \textbf{LIME}~\cite{ribeiro2016should}.
In LIME, the neighbourhood is obtained by sampling instances from a Gaussian distribution centered on $\boldsymbol{z}_e$.
Locality is introduced by considering weights derived from the exponentiated negative euclidean distance between these instances and $\boldsymbol{z}_e$.
The default bandwidth (or kernel width) is a heuristic based on the dimension of the input space, hence failing to adapt to every unique case and its possible intricacies.
The consequence of defining locality this way is the risk of LIME not properly capturing the relevant local classification behaviour of the black-box~\cite{laugel2018b}. 
The weighted neighbourhood is then used to train a linear decision model. 
LIME does not explore the decision boundary directly, but instead samples the dataset, such that it is close (in Euclidean distance) to the instance in question. 





To circumvent LIME's issues, \textbf{GSLS}~\cite{laugel2018b} exploits the structure of the decision boundary locally, by first finding the nearest point of the opposite class, and then generating instances uniformly within a hyper-ball in its vicinity.
As opposed to LIME, GSLS clearly defines which part of the decision boundary is relevant: the nearest point on the decision boundary.
An issue is how to define how large the hyper-ball neighbourhoods should be. 
Furthermore, the decision boundary may require more than one point on the decision boundary in several directions to be fully captured by the neighbourhood.





\textbf{LORE} (Local Rule-Based Explanations) \cite{guidotti2018local} also makes use of the decision boundary in the vicinity the instance to be explained to construct the neighbourhood.
The construction of the neighbourhood is posed as a set of optimisation problems: one for finding instances of the same class as $\boldsymbol{z}_e$, and one for finding instances not belonging to that class.
A genetic algorithm is used to maximise the objective function.
As part of the optimisation problem, a distance function that is a mixture of euclidean distance, for continuous features, and simple feature matching, for discrete features, is considered.
This allows the approach to detect the variations of the local decision boundary.
In comparison with GSLS, LORE goes a step further and aims at an in-depth exploration of the decision boundary surrounding the instance.
While they also make use of the decision boundary, and explore beyond the nearest point, there is not a clear understanding of what ``local'' means in this case. 
A different direction in defining the neighbourhood is based on relying on the training instances to favour the most important features.
As such, \textbf{LEAP} (Local Embedding Aided Perturbation) \cite{localembeddings} tries to extract the relevant subspace for the instance by employing dimensionality reduction techniques at a local level.
They use the \textit{Local Intrinsic Dimensionality} (LID) \cite{localembeddings} to identify the dimensionality of the subspace at the vicinity of the instance.
Once the $LID$ is identified, a dimensionality reduction technique is used (e.g., Principal Components Analysis (PCA)).
$\boldsymbol{z}_e$ is then projected in the new subspace giving $\hat{z}_e$.
The neighbourhood and associated weights are then respectively sampled and computed similarly to LIME.
The elements of the neighbourhood are then mapped back to the original feature space, where the surrogate is fitted.
LEAP first tries to identify the local manifold; then it proceeds in a similar fashion to LIME, but in this sub-space of reduced dimensionality.

In \textbf{PALEX} \cite{jia2019palex} the neighbourhood is constructed by repeating the procedure of: (1) randomly selecting a subset of features from $\boldsymbol{z}_e$, and (2) replacing them with values from $\boldsymbol{X}$.
The calculation of the weights relies on a set of frequent patterns extracted from the training set.
The distance depends on how often two instances appear together on frequent patterns. 
The main difference of PALEX with previously mentioned methods is the weight function that utilises pattern mining.
This type of weighting function goes beyond the euclidean measure and aims at identifying manifolds in the global data structure.





An alternative strategy is followed by \textbf{KernelSHAP}~\cite{Lundberg_Lee_2017} which is motivated by the theory of Shapley Values from cooperative game theory.
The relevant (for us) contribution of the paper is to pose Shapley Values as the solution of a weighted linear regression problem when the design matrix ($X_{shap}$), weights and targets ($y$) are carefully chosen.
The design matrix, $X_{shap} \in \mathbb{R}^{2^d\times d}$, is the set of all binary vectors of size $d$ (the number of features), it represents all the subsets of the powerset of features.
The weight matrix is a diagonal matrix with the Shapley kernel weights for each row of the design matrix.
Each row of $X_{shap}$ represents two sets of features: the clamped $Q \in [d]$ (features in the current subset) and those that will not be clamped $\neg Q$ (features that are not in the current subset).
With this in mind, we represent with $y_Q$ the target corresponding to the row of $X_{shap}$ that has features $Q$ clamped.
We have in theory $y_Q = \mathbb{E}[f(Z)~|~Z_i=z_{e,i},~\forall i \in Q]$, which is understood as the expected output of the black-box if we fix the features in set $Q$ according to the values in $z_e$.
The exact computation of this is very computationally expensive.
Hence, the authors resort to approximations (illustrated in Eqs.9-12 of \cite{Lundberg_Lee_2017}) which include the assumptions of feature independence and model linearity for the black-box.
As part of the approximations, information needs to be extracted from the black-box to compute the expected outputs, which we present in Fig.~\ref{fig:sota_neighbourhoods}.
Also, the high-dimensionality of $X_{shap}$ poses further computational issues, which require additional relaxations of the original formulation, such as feature selection procedures, or added regularisation schemes.
If formalising the problem of explaining a prediction with Shapley Values from cooperative game theory fits the explanation needs, KernelSHAP proposes an interesting approach.
However, many assumptions and approximations are necessary to make it work in practice, raising questions on the meaning of the generated output, in particular how they compare to the true Shapley Values.

As we have shown in Section~\ref{sec:background} the approaches have a common process for the construction of a local surrogate: (1) neighbourhood sampling, and (2) model fitting.
On the other hand, we made apparent in Section~\ref{sec:comparison} that there is a lack of consensus on what properties these models should possess.
Table~\ref{tab:comparison} provides a comparison of the different neighbourhood strategies of the approaches discussed in this section.
There is a general agreement that the surrogate should reflect the behaviour of the black-box in the locality of the instance to be explained, but there is no consensus on what this means.
As is illustrated in Figure~\ref{fig:sota_neighbourhoods}, even though the approaches share the same goal (explaining a prediction) they extract different information from the black-box model.

Furthermore, the explanation provided by an approach cannot be traced back to an explanation need, because it is not a well-defined problem (not explicitly defined, heuristic or approximations with unknown consequences on the initial promise).
Thus, there is no real understanding of what the explanation means, preventing proper comparisons between approaches.
This impacts the research in the field and the application of explainability in practice.
We elaborate on this issue in the following section.

\section{Discussion}

\begin{figure}
    \centering
    \includegraphics[width=0.95\columnwidth]{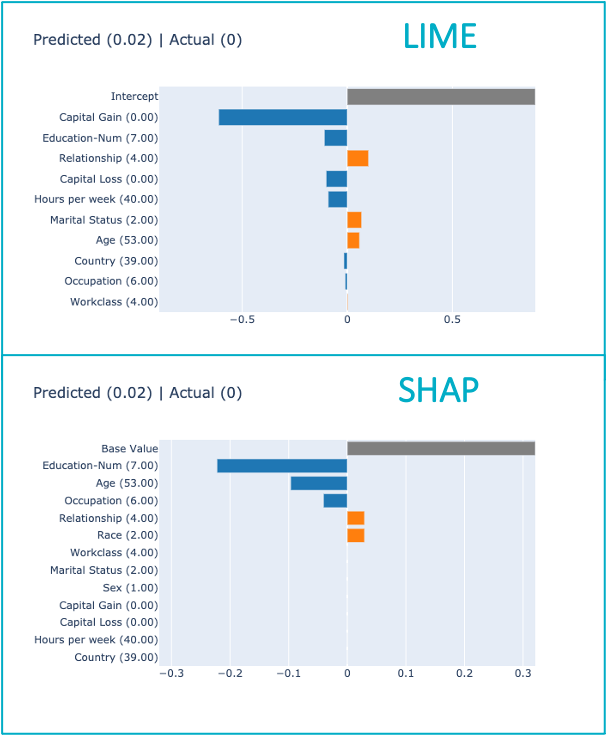}
    \caption[adult dataset]{Explanations returned by LIME and SHAP (with base parameter values) for a prediction made by a Random Forest Classifier for a randomly picked instance from the Adult dataset}
    \label{fig:LIME_vs_SHAP_explanations}
\end{figure}

In the previous section we saw that approaches under the same category of local-surrogate explainers, have different objectives and capture different information from the black-box.
While this study has been focusing on the neighbourhood, our observations on the diversity of the information captured by local surrogate methods can be used to imply a diversity in the explanations produced.
As illustrated in Figure~\ref{fig:LIME_vs_SHAP_explanations}, the same instance on Adult dataset~\footnote{https://archive.ics.uci.edu/ml/datasets/adult} presents conflicting explanations generated by KernelSHAP and LIME.
A practitioner is confronted with conflicting explanations with no way of comparing, or evaluating the methods.

It appears the problem of explaining black-box prediction has been tackled backwards by proposing solutions first, before defining the problem itself.
As a consequence, as shown by studies, interpretability approaches tend to be misunderstood and followed blindly~\cite{Kaur2020}.
For instance, the relevance of Shapley values to explain individual predictions has been questioned recently~\cite{kumar2020problems,weerts2019human}).
This did not prevent some approaches such as KernelSHAP to be extensively used,
in spite of the issues mentioned in this paper, or in other studies~\cite{kumar2020problems}.

The machine learning community should pay more attention to user-centric works that attempt to specify interpretability requirements in link with user needs~\cite{miller2019explanation,liao2020questioning}).
From these specifications alone should arise proper interpretability objectives to compare or design efficient interpretability approaches.

\bibliography{main}
\bibliographystyle{icml2021}

\end{document}